\def\year{2019}\relax
\documentclass[letterpaper]{article} %
\usepackage{aaai19}  %
\usepackage{times}  %
\usepackage{helvet} %
\usepackage{courier}  %
\usepackage[hyphens]{url}  %
\usepackage{graphicx} %
\usepackage{graphicx}  %
\usepackage{tikz}
\usepackage{times}
\usepackage{soul}
\usepackage{url}
\usepackage[utf8]{inputenc}
\usepackage{amsmath}
\usepackage{amssymb}
\usepackage{amsthm}
\usepackage{booktabs}
\usepackage[linesnumbered,algoruled,boxed,lined]{algorithm2e}
\usepackage[skip=0cm]{subcaption}
\usetikzlibrary{positioning}

\newtheorem{observation}{Observation}

\newtheorem{example}{Example}

\title{On Hard Exploration for Reinforcement Learning:\\ A Case Study in Pommerman}
\author{Chao Gao\textsuperscript{\rm 1}\thanks{Work performed as a part-time intern at Borealis AI.} \Large \textbf{ Bilal Kartal\textsuperscript{\rm 1}, Pablo Hernandez-Leal\textsuperscript{\rm 1} and Matthew E. Taylor\textsuperscript{\rm 1}}\\ %
\textsuperscript{\rm 1}Borealis AI\\ %
Edmonton Canada\\
cgao3@ualberta.ca, \{bilal.kartal, pablo.hernandez, matthew.taylor\}@borealisai.com %
}
\begin{document}
\maketitle
\begin{abstract}
How to best explore in domains with sparse, delayed, and deceptive rewards is an important open problem for reinforcement learning (RL). This paper considers one such domain, the recently-proposed multi-agent benchmark of Pommerman. This domain is very challenging for RL --- past work has shown that model-free RL algorithms fail to achieve significant learning without artificially reducing the environment's complexity. In this paper, we illuminate reasons behind this failure by providing a thorough analysis on the hardness of random exploration in Pommerman. While model-free random exploration is typically futile, we develop a model-based automatic reasoning module that can be used for safer exploration by pruning actions that will surely lead the agent to death. We empirically demonstrate that this module can significantly improve learning.
\end{abstract}

Efficient exploration is a long-standing problem in reinforcement learning~\cite{haarnoja2017reinforcement,nachum2016improving}. Thrun~(\citeyear{thrun1992cient}) divided exploration methods in two: undirected, which are closely related to a random walk with the advantage of not using specific knowledge~\cite{szepesvari2010algorithms,still2012information}; and directed, which need domain specific knowledge with the advantage of generally better exploration. A subcategory of directed exploration are model-based approaches, which usually take into account how often a state-action pair has been visited~\cite{brafman2002r,strehl2008analysis}. However, count approaches suffer in non-tabular cases with large state spaces.
In these challenging cases, the agent is usually equipped with a function approximator for effective generalization across familiar and unfamiliar states.
An example that extends tabular count-based exploration to non-linear function approximation is pseudo-count~\cite{bellemare2016unifying}, which can be used to evaluate states that have not been visited in the past.

However, these methods usually struggle in hard-exploration problems~\cite{ecoffet2019go} (e.g., Atari games such as Montezuma's Revenge and Pitfall) where many long sequences of actions are needed before obtaining obtaining non-zero external signals.

\begin{figure}
\centering
\includegraphics[scale=0.24]{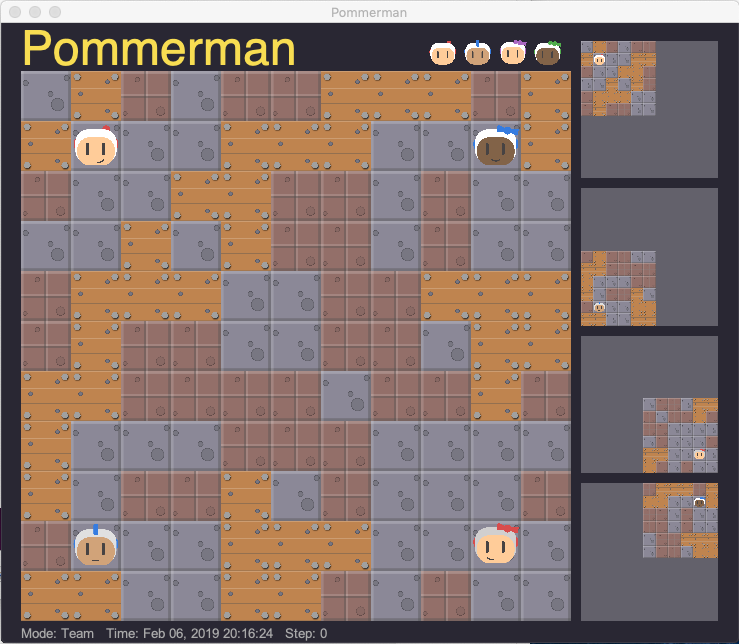}
\caption{The game of Pommerman in Team mode, where each two-diagonal agents form a team. The right column shows the partial view of the board for each agent; the view is fully observable in FFA mode.} \label{fig:pommerman}
\end{figure}

This paper considers the exploration problems that arise in a recently proposed benchmark for (multi-agent) reinforcement learning: Pommerman~\cite{resnick2018pommerman}. The environment is based on the classic console game \emph{Bomberman}
that involves 4 bomber agents initially placed at the four corners of a board. The game can be played in Free-For-All (FFA) or Team modes. In FFA the winner is the last agent that survives; in Team mode each two diagonal agents form a team, and one team wins if it successfully destroys all members in the other team. In either mode, at every step, each agent issues an action simultaneously from 6 discrete candidate actions: moving \verb|left|, \verb|right|, \verb|up|, \verb|down|,  placing a \verb|bomb|, or \verb|stop|.
The \verb|bomb| action is legal as long as the agent's \verb|ammo| is greater than 0, and any illegal action (such as moving towards a wall) is superseded with \verb|stop| by the environment. Figure~\ref{fig:pommerman} shows a snapshot of the game.

Pommerman is challenging for multi-agent learning, in particular, model-free reinforcement learning, predominantly because exploration is difficult due to:
(1) delayed action effects: the only way to make a change to the environment (e.g., kill an agent) is by means of bomb placement, but the effect of such an action is only effective when the bomb explodes after 10 time steps;
and (2) sparse and deceptive rewards: the former refers to the fact that the only non-zero reward is obtained at the end of an episode. The latter refers to the fact that quite often a winning reward is due to the opponents' involuntary {suicide}, which makes reinforcing an agent's action based on such a reward \emph{deceptive}.

While techniques like Reward Shaping and Difference Rewards~\cite{agogino2008analyzing,devlin2014potential} have been used to deal with (2), for (1), efficient and safe exploration is crucial because failing to avoid the side-effect of the bomb actions results in irreversible events~\cite{farquhar2017treeqn,racaniere2017imagination}, i.e., the agent is eliminated.
Although it has been suggested that model-free RL does not learn well in Pommerman, particularly because the exploration with bomb action is highly correlated to losing~\cite{resnick2018pommerman,resnick2018backplay,kartal2018using}, a formal analysis about the exploration difficulty is lacking.
In this paper we fill this gap by presenting an analysis on the exploration hardness in Pommerman: we show that suicides happen frequently during learning because of the nature of using model-blind random actions for exploration. Furthermore, the high rate of suicides has a direct effect on the samples needed to learn. We exemplify this in a case for which an \emph{exponential} number of samples are needed to obtain a positive experience. This highlights that performing non-suicidal bomb placement could require complicated, long-term, and accurate planning.
We  then propose an efficient reasoning module that prunes unsafe actions. We experimentally demonstrate the usefulness of the module by comparing the learning performances with and without such a module, and further show the strength of obtained players by competing against \texttt{SimpleAgent}, the official baseline provided by Pommerman %

\section{Analysis of Random Exploration}
\label{sec:analysis}
Before presenting our analysis in Pommerman, we first describe a related classic problem of random walk in two-dimensional world, then we generalize to the case of random walk in a world with \emph{obstacles}, and finally we describe the results for Pommerman.

\subsection{Obstacle-free 2D grid world}
We start with a simplified scenario where an agent sits on a 2d world and has five actions --- \texttt{stop, left, right, up, down} --- to navigate around. Assume the agent's original location is marked as $(0,0)$ and follows an uniform random policy.
The question of interest is: at time step $t \geq 0$, what is the probability that the agent will be at cell $(i,j)$ for arbitrary $(i,j)$?

Each step the agent has five actions to take, thus after $t$ steps, there are in total of $5^t$ ``paths,'' because the probability of each path is the same, the probability that the agent be at position $p=(i,j)$ can be computed as $\frac{N(p, t)}{5^t}$, where $N(p, t)$ is the number of paths that lead the agent to position $p$ at time $t$. Therefore, %
the  question is how to calculate $N(p,t)$ for any position $p=(i,j)$ and time step $t$.

After any tick the agent can at most move to a cell that is one step away. Therefore, to arrive at $p$ at time step $t$, the agent must get to the neighboring positions within $t-1$, as shown by the recursive relation:

\begin{equation}
    N(p,t) = \begin{cases}
    \sum_{q \in \text{neighbor}(p)}^{ } & N(q, t-1) +  N(p, t-1),  \\
    1, & p=(0,0) \text{ and } t=0
    \end{cases}
    \label{eq:freeworld}
\end{equation}
\noindent where \emph{neighbor}$(p)=\{(i-1,j),(i,j-1),(i+1,j),(i,j+1)\}$, given $p=(i,j)$.

This recursion
expresses that to be on position $(i,j)$ at time $t$, in the previous time step $t-1$, the agent must be at $(i-1,j), (i,j-1), (i+1,j), (i,j+1)$ or $(i,j)$ from where one  action can be taken by the agent, resulting in $(i,j)$.

\begin{figure}[t]
    \centering
    \begin{subfigure}[b]{0.32\linewidth}
    \centering
    \begin{tikzpicture}
    \draw[step=0.5cm] (0.25,0.25) grid (2.25,2.25);
    \draw (1.25,1.25) node {$1$};
    \end{tikzpicture} \\
    \caption*{t=0}
    \end{subfigure}
\begin{subfigure}[b]{0.32\linewidth}
    \centering
    \begin{tikzpicture}
    \draw[step=0.5cm] (0.25,0.25) grid (2.25,2.25);
    \draw (1.25,1.25) node {1};
    \draw (0.75,1.25) node {1};
    \draw (1.25,0.75) node {1};
    \draw (1.75,1.25) node {1};
    \draw (1.25, 1.75) node {1};
    \end{tikzpicture} \\
    \caption*{t=1}
\end{subfigure}
\begin{subfigure}[b]{0.32\linewidth}

    \centering
    \begin{tikzpicture}
    \draw[step=0.5cm] (0.25,0.25) grid (2.25,2.25);
    \draw (1.25,1.25) node {5}; %
    \draw (0.75,1.25) node {2};
    \draw (1.25,0.75) node {2};
    \draw (1.75,1.25) node {2};
    \draw (1.25, 1.75) node {2};
    \draw (2.25, 1.25) node {1};
    \draw (1.25, 2.25) node {1};
    \draw (0.25, 1.25) node {1};
    \draw (1.25, 0.25) node {1};
    \draw (1.75, 1.75) node {2};
    \draw (0.75, 1.75) node {2};
    \draw (0.75, 0.75) node {2};
    \draw (1.75, 0.75) node {2};
    \end{tikzpicture} \\
    \caption*{t=2}
\end{subfigure}
    \caption{At step 0, the agent is located at the origin; at step 1, the agent could be at any position 1-step away; at step 2, the agent can at most reach a position that is two-steps away. The probability of being at each position can be computed by dividing the number of all feasible combinatorial trajectories, i.e., $5^t$.}
    \label{fig:freeworld}
\end{figure}
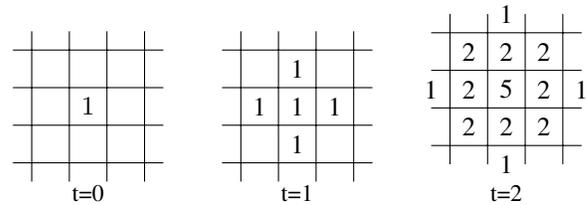

Since $N(p,t)$ in Equation~\ref{eq:freeworld} depends solely on the information in time $t-1$, and solution to the base case ($t=0$) is given, this indicates that $\forall p, t\geq 0, N(p,t)$ can be computed bottom-up using dynamic programming by iteratively increasing the value of $t$. As an illustration, Figure~\ref{fig:freeworld} shows the number of paths to each position for time steps $t=0,1,2$, where unmarked positions are out of reach of the agent --- they can also be filled with 0. Note also that each number in a sub-figure can be computed summing all the numbers of its neighbors and itself from the sub-figure on the left.

\subsection{Generalization to arbitrary obstacles}
Now we generalize to a scenario where: (i) the board is bounded and trying to move outside results in \emph{stop}; and (ii) there are arbitrary static obstacles that the agent cannot pass through, i.e., the agent stays in the same position. This scenario is similar to a typical situation in Pommerman, where an agent is positioned on a board with limited size, has \texttt{ammo} of 0 (no bomb placement allowed) and can use the remaining five legal actions to explore the world.\footnote{Note that stripping the invalid \texttt{bomb} action when \texttt{ammo=0} does not violate the model-free principle, since from the egocentric perspective of view, \texttt{ammo} is an attribute of the agent rather than a specification from model of the environment.}

Only a minor revision to Equation~\ref{eq:freeworld} is needed to represent the number of paths of being at each $p=(i,j)$ after $t$ steps, given a board configuration $\mathcal{B}$, as:
\begin{equation}
    N(p, t, \mathcal{B}) =  \sum_{a \in A} N(prev(a, p, \mathcal{B}), t-1, \mathcal{B}), \quad t>0
\end{equation} \label{eq:N_obstacle}
\begin{equation}
   N(p, t, \mathcal{B}) =  1, p=(0,0) \wedge t=0
\end{equation} \label{eq:N_obstacle_base}

\noindent
where $A=\{$\texttt{stop, left, right, up, down}$\}$ is the action set and $prev(a, p, \mathcal{B})$ returns the previous position the agent must be if action $a$ leads the agent to position $p=(i,j)$ on a board configuration $\mathcal{B}$. %

\begin{figure}
    \centering
    \begin{subfigure}[b]{0.32\linewidth}
    \centering
    \begin{tikzpicture}
    \draw[step=0.5cm] (0,0) grid (2.5,2.5);
    \draw (1.25,1.25) node {$1$};
    \draw (1.25, 0.75) node {$\blacksquare$};
    \end{tikzpicture} \\
    \caption*{t=0}
    \end{subfigure}
\begin{subfigure}[b]{0.32\linewidth}
    \centering
    \begin{tikzpicture}
    \draw[step=0.5cm] (0,0) grid (2.5,2.5);
    \draw (1.25,1.25) node {2};
    \draw (0.75,1.25) node {1};
    \draw (1.25,0.75) node {$\blacksquare$};
    \draw (1.75,1.25) node {1};
    \draw (1.25, 1.75) node {1};
    \end{tikzpicture} \\
    \caption*{t=1}
    \end{subfigure}
\begin{subfigure}[b]{0.32\linewidth}
    \centering
    \begin{tikzpicture}
    \draw[step=0.5cm] (0,0) grid (2.5,2.5);
    \draw (1.25,1.25) node {7}; %
    \draw (0.75,1.25) node {3};
    \draw (1.25,0.75) node {$\blacksquare$ };
    \draw (1.75,1.25) node {3};
    \draw (1.25, 1.75) node {3};
    \draw (2.25, 1.25) node {1};
    \draw (1.25, 2.25) node {1};
    \draw (0.25, 1.25) node {1};
    \draw (1.75, 1.75) node {2};
    \draw (0.75, 1.75) node {2};
    \draw (0.75, 0.75) node {1};
    \draw (1.75, 0.75) node {1};
    \end{tikzpicture} \\
    \caption*{t=2}
    \end{subfigure}
    \caption{At step 0, the agent is located at the origin and there is one obstacle in the board represented by $\blacksquare$; step 1, the agent could be at any position 1-step away; at step 2, the agent can at most reach to a position that is two-steps away. The probability of being at each position can be computed by dividing the corresponding normalization constant $5^t$.}
    \label{fig:pommermanworld}
\end{figure}
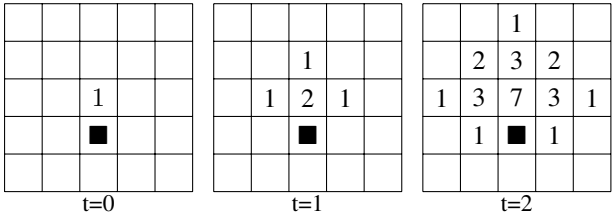
Figure~\ref{fig:pommermanworld} shows an example for $t=0,1,2$. In this case, there is one obstacle in the board (represented by $\blacksquare$).

\subsection{Suicide study: Pommerman}
In Pommerman, it has been noted that the action of bomb placing is highly correlated to losing~\cite{resnick2018pommerman}, which is presumably the major impediment for achieving good results using model-free reinforcement learning. Now, we provide a formal analysis of the suicide problem that was suggested to be the reason to delay or prevent the agent from learning the \emph{bombing skill}~\cite{kartal2018using} when an agent follows a model-blind, random exploration policy.

In Pommerman, an agent can only be killed when it intersects with an exploding bomb's flames, then we say a \emph{suicide} event happens if the death of the agent is caused by its own bomb. For the ease of exposition we consider the following simplified scenario: (i) the agent has \texttt{ammo=1} and has just placed a bomb (i.e., the agent now sits on the bomb with \texttt{ammo=0}); (ii) for the next time steps until the bomb explodes, the agent has $5$ actions available at every time step; and (iii) other items on the board are static.

One difference between this case and the previous section is that the agent sits on an ``obstacle'' (i.e., a bomb), and once it moves off it cannot come back. %
This is in contrast to the previous assumption where the number of paths to any position is always related to its adjacent cells. Formally, the original position $p$ of the agent represents a singleton point that if the agent is at one neighbor of $p$ in time $t$, it cannot go back to $p$ in any $t'\geq t+1$. Because of this special point, we have the following observations:

\begin{observation}
Assuming $p$ is the origin and the bomb has not exploded yet, the number of paths of being at $p$ after $t$ time steps on a board $\mathcal{B}$ is given by:
$$N(p, t, \mathcal{B}) = (|obstacle(p, \mathcal{B})|+1)^t \quad\text{with}\quad \forall t \geq 0$$
where $obstacle(p, \mathcal{B})$ is the set of obstacle cells adjacent to $p$.
\end{observation} \label{observ:1}

\begin{proof}
To stay at the origin, at each step, the agent must take an action either is \texttt{stop} or results in a bounce back, so that after $t$ steps, there is in total $(|obstacle(p, \mathcal{B})|+1)^t$ possible ``paths.'' Note that $|obstacle(p, \mathcal{B})| \leq 4$.%
\end{proof}
\noindent Due to the special treatment of the origin, the asymmetric impact needs to be propagated to its adjacent neighbours by the following initialization (noted as $\hat{N}$):

\begin{observation}
For any passable neighbor $q \in neighbor(p)$, after $t$ time steps in $\mathcal{B}$:
$$\hat{N}(q, t, \mathcal{B}) = \hat{N}(p, t-1, \mathcal{B}) $$ where $p$ is the origin and
$\hat{N}(p, t-1, \mathcal{B}) = (|obstacle(p, \mathcal{B})|+1)^{t-1}$.
\end{observation} \label{observ:2}
\noindent The other cells can be initialized $\hat{N}$ with 0, {because they are out of direct influence of the origin.} %

Now, the remaining problem can be solved exactly by applying the scheme in the previous section, except that the initialization value has to be added to the recursion:
\begin{equation}
    N(p, t, \mathcal{B}) = \hat{N}(p, t, \mathcal{B}) + \sum_{a \in A} N(prev(a, p, \mathcal{B}), t-1, \mathcal{B}) , ~t \geq 0
\end{equation} \label{eq:pommerman}

Now, we use the previous analysis to exemplify a random, best, and worst cases that happen in Pommerman. For brevity, we call the number of paths to a cell $c$ as the \emph{N value of} $c$, or \emph{N value of} $c$ at time $t$ if $t$ has to be explicitly noted.

\begin{example}
We take a random starting board configuration, depicted in Figure~\ref{fig:pommerman}, and compute the $N$ values ending at each position for $t=0$ and $t=3$ in  Figure~\ref{fig:pommerman_t0}. This is a typical starting board in Pommerman, where every agent stays at its corner; they are disconnected with each other by randomly generated obstacles. One noticeable phenomenon is that the $N$ values are concentrated around its origin.

We also compute for $t=9$ (time steps needed for a bomb to explode) before and after normalizing the number of paths into probabilities in Figure~\ref{fig:pommerman_prob9}. Note that even when the configuration of the agents is slightly different their probabilities of ending up with suicide are $\approx 40\%$ ($0.39, 0.38, 0.46, 0.38$, counterclockwise starting from upper-left corner).

\begin{figure}[t]
    \centering
    \begin{subfigure}{0.49\linewidth}
    \centering
    \includegraphics[scale=0.84]{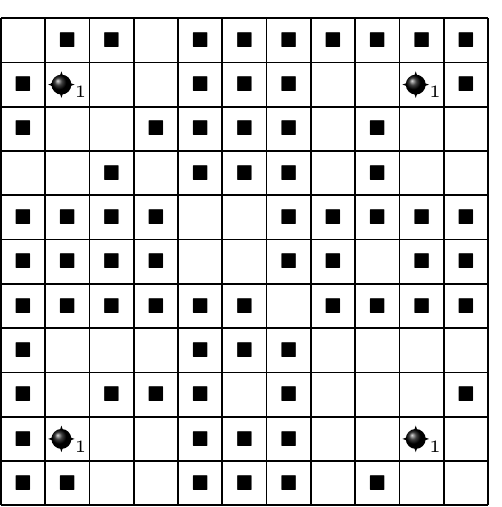}
    \caption*{t=0}
    \end{subfigure}
    \begin{subfigure}{0.49\linewidth}
    \centering
    \includegraphics[scale=0.84]{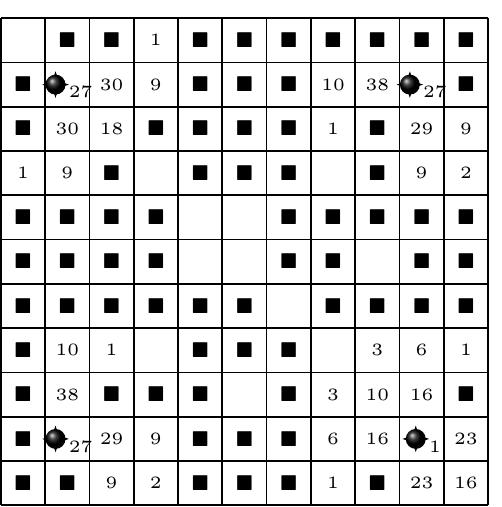}
    \caption*{t=3}
    \end{subfigure}
    \caption{In Pommerman, after placing a bomb, each agent sits on the bomb; left and right figures are the number of paths of being at each cell at step 0 and 3, respectively. Probabilities can be obtained by dividing by $5^t$.}
    \label{fig:pommerman_t0}
\end{figure}

\begin{figure}[t]
    \centering
    \begin{subfigure}{0.47\linewidth}
        \centering
        \includegraphics[scale=0.185]{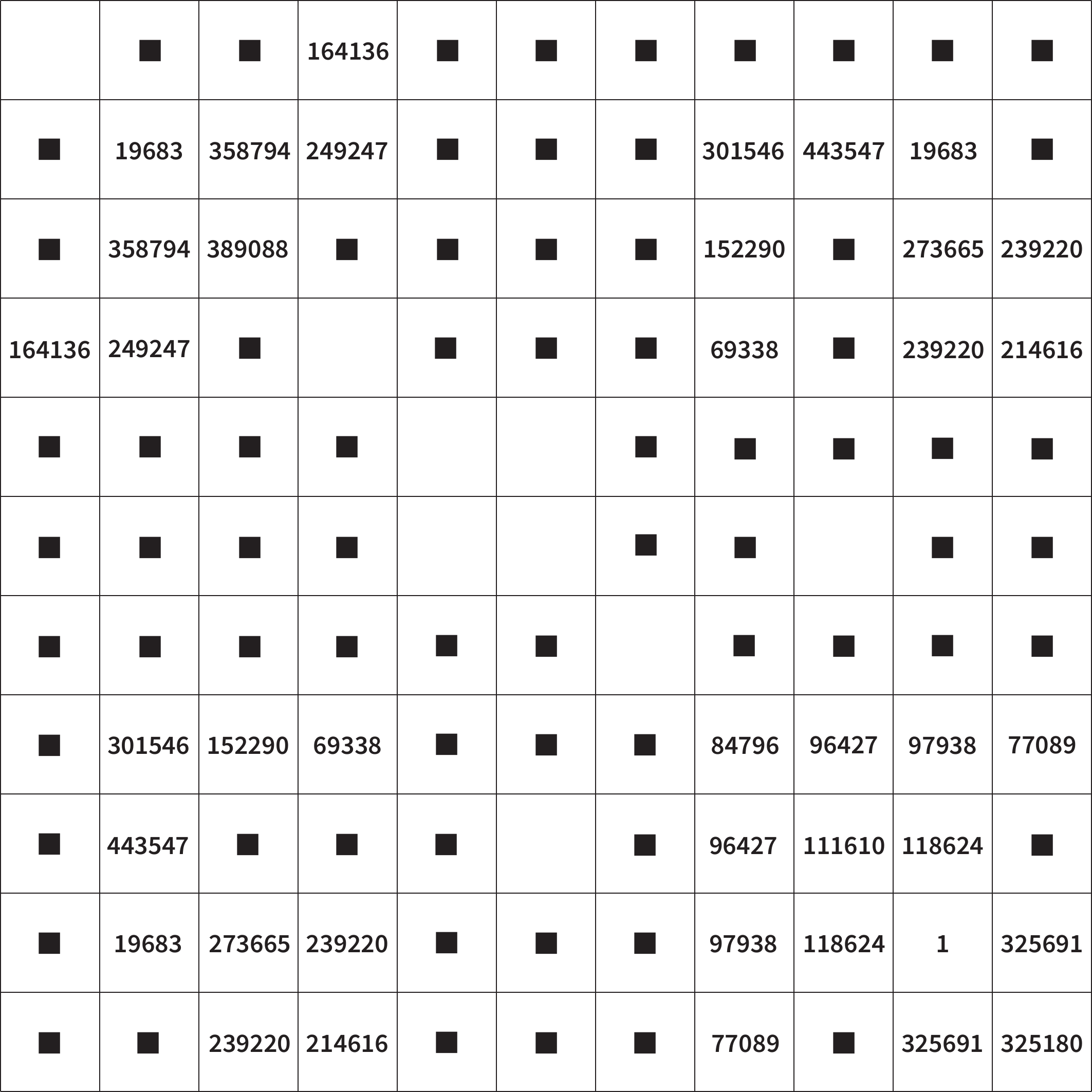}
        \\
        \caption*{t=9, $N$ values}
    \end{subfigure}
    \quad
    \begin{subfigure}{0.47\linewidth}
        \centering
        \includegraphics[scale=0.8]{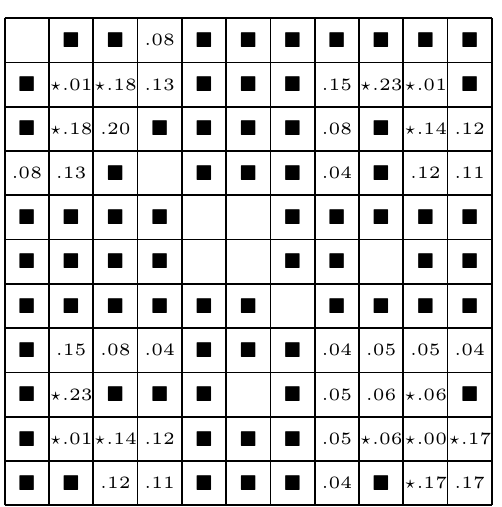}
        \\
        \caption*{t=9, probabilities}
    \end{subfigure}
    \caption{After 9 steps, the $N$ value and probability of being at each cell for each agent. $\star$ indicates the cell is covered by flames.}
    \label{fig:pommerman_prob9}
\end{figure}
\end{example}

Indeed, the problem of suicide stems from acting randomly without considering constraints of the environment. In extreme cases, in each time step, the agent may have only one survival action, which means the safe path is always unique as $t$ increases while the total number of {feasible} paths {the agent can traverse}
grows exponentially. We illustrate this by the following corridor example.

\begin{example}
Figure~\ref{fig:corridor} shows a worst-case example: the agent is in a corridor formed by wood at two sides and places a bomb. If using model-blind exploration the chance of suicide is extremely high since among the $5^9$ ``paths,'' \emph{only one} of them is safe. In order to survive, it must precisely follow the right action at each time step. %
This also implies that for sub-problems of such in Pommerman, to acquire one positive behaviour example requires \emph{exponential} number of samples.
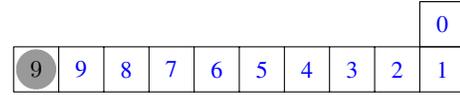
\begin{figure} \small
    \centering
    \begin{tikzpicture}[scale=0.6]
    \draw (0,0) grid (10, 1);
    \draw (9,1) rectangle (10,2);
    \draw (0.5,0.5) node[shape=circle, opacity=0.4, fill=black, text opacity=1, draw, inner sep=0pt] {${~~9~~}$};
    \foreach \x in {1,...,9}{
        \pgfmathtruncatemacro{\label}{10-\x}
        \draw (\x+0.5,0.5) node[shape=circle, color=blue] {\label};
    };
    \draw (9.5,1.5) node[shape=circle, color=blue] {0};
    \end{tikzpicture}
    \caption{The corridor scenario: the agent places a bomb with \texttt{strength=10} on the leftmost cell. For each passage cell, the marked value means the minimum number of steps it is required to safely evade from impact of the bomb. After placing the bomb, in the next step the bomb has life of $9$, thus in the remaining $9$ ticks, the agent must exactly take the right action to evade. }
    \label{fig:corridor}
\end{figure}

\end{example}

The previous example shows a worst-case scenario, now we present also the best-case. \begin{observation}
Suppose the agent starts by sitting on a bomb and the agent uniform-randomly walks by taking actions \texttt{left},  \texttt{right},  \texttt{up},  \texttt{down}, or \texttt{stop}. The best board configuration for the agent not committing suicide is when there are no obstacles (except the bomb at the origin) and the agent is at the center of the board.
\end{observation}

\begin{figure}
    \centering
    \begin{subfigure}{0.47\linewidth}
        \centering
        \includegraphics[scale=0.8]{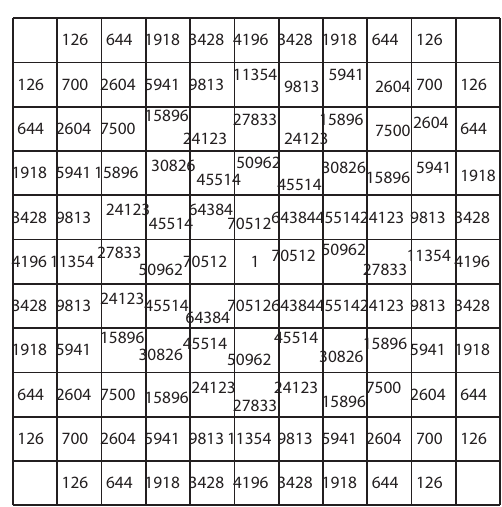}
        \\
        \caption*{t=9, $N$ values}
    \end{subfigure}
    \quad
    \begin{subfigure}{0.47\linewidth}
    \centering
    \includegraphics[scale=0.8]{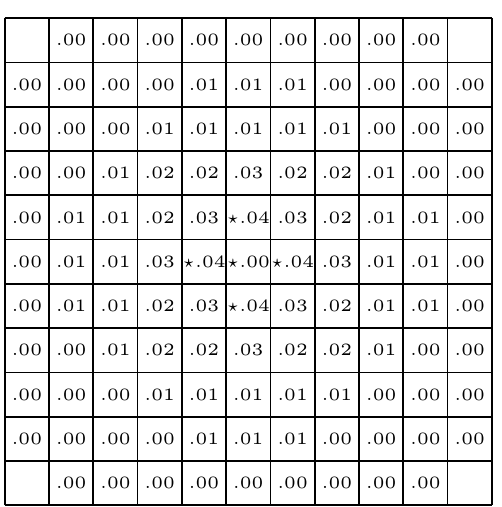}
    \\
    \caption*{t=9, probabilities}
    \end{subfigure}
    \caption{Best case for an agent to be after placing a bomb after 9 steps: empty board. The $N$ value and probability of being at each cell for each agent; $\star$ indicates the cell is covered by flames. }
    \label{fig:best_case}
\end{figure}

Figure~\ref{fig:best_case} shows the computation results in detail for this best-case scenario where the suicide probability is $0.16$. However, in Pommerman, such an ideal case does not exist at all --- we have seen from Figures~\ref{fig:pommermanworld},~\ref{fig:pommerman_t0},~\ref{fig:pommerman_prob9} and~\ref{fig:corridor}, usually the probability is higher. Specifically, for a whole game, if $k$ bombs have to be placed, suppose the agent's suicide probability is $x$ each time, then the suicide rate is thus $1-(1-x)^k$ which increases as $k$ increases --- this puts the agent in complicated scenario: placing more bombs is required for playing well while at the same time it also results more probable suicide. For example, suppose an agent has to survive $k=8$ bombs to carve out a path to engage with enemies (not an uncommon scenario in Pommerman), then assuming $x= 0.4$, the probability of death after $8$ bombs is $ \approx 1$.  Furthermore, even the agent somehow managed to survive from these bombs, no reward is observed as the positive reward signal appears only when all opponents are dead

\section{Action Pruning}
\label{sec:actionFilter}
In this section we show that if the model-free demand is relaxed, an action pruning algorithm for safe exploration exists in Pommerman.

In the case illustrated in Figure~\ref{fig:corridor}, we see that the essence for survival is to try to navigate to a position that is free from the threat of future bombs (rightmost upper cell in Figure~\ref{fig:corridor}). Let us call the \emph{safety value} of a passable cell the minimum number of steps to reach a safe region (that is out of the reach of any bombs). Assuming static bombs, %
we can prune unsafe actions by comparing the post-action location's \emph{safety value} and its minimum bomb life value. Figure~\ref{fig:corridor} shows the safety values for the corridor example.

This idea can be generalized to arbitrary bombs: given a board situation and the agent's position, each resulting cell after taking an action can be computed --- the remaining problem is to compute its \emph{safety value} as well as the minimum life value among the bombs covering this position, where we say a position is \emph{covered by a bomb} if (after the explosion) the position is within reach of the flames.

We describe the complete algorithm for computing \emph{safety values} in Algorithm~\ref{algo:pruning} (\texttt{MinEvadeStep} function). The input to the algorithm are attributes of the agent as well as the board information that the agent can see. The output (calling \texttt{GetSafeActions}) is set of \emph{safe} actions. Exact computation is impossible for partially observable environments, but can be approximated by treating unseen areas as walls.

\begin{algorithm}[hpt]
 \footnotesize
  \DontPrintSemicolon
  \KwIn{$board, agent, A$}
  \KwOut{$\hat{A}$}
  \SetKwFunction{FMain}{MinEvadeStep}
  \SetKwProg{Fn}{Function}{:}{}
   \Fn{\FMain{$board, p, history$}}{
        $u,l \gets \text{\texttt{FindMaxMinBombCovering}}(board, p)$ \;
         \uIf{$u = - \infty $}{
             \tcc{no bomb covering $p$}
             \Return $0$ \;
         } \uElseIf{$|history| \geq u$}{
             \tcc{bomb would have exploded upon arrival}
             \uIf{$|history| > u + 2 $}{
             \tcc{2 is flame life}
                 \Return $0$ \;
             } \uElse{
                 \Return $\infty$\;
             }
         } \uElseIf{$|history| \geq l$}{
             \uIf{$|history| > l + 2 $}{
                 \Return $0$ \;
             }\uElse{
                 \Return $\infty$\;
             }
         }
         $num \gets \infty$ \;
         \For{$a \in \{left, right, up, down\}$}{
             $q \gets \text{\texttt{NextPosition}}(p, a)$\;
             \uIf{$q \not \in history$}{
                 $num \gets \min(num, 1+\text{\texttt{MinEvadeStep}}(board,q, history\cup\{q\}))$\;
             }
         }
         \Return $num$\;
  }
  \SetKwFunction{FMain}{GetSafeActions}
  \Fn{\FMain{$board, agent, A$}}{
        $\hat{A} \gets \emptyset$\;
        Let $p$ be agent's position\;
        \For{$a \in A=\{left, right, up, down, stop, bomb\}$}{
            $board, q \gets \text{\texttt{Next}}(board, p,a)$ \;
            \uIf{$a = stop$ or $a=bomb$}{
                $H \gets \{None, q\}$\;
            }\uElse{
            $H \gets \{q\}$ \;
            }
            $n \gets \text{\texttt{MinEvadeStep}}(board,q, H)$ \;
            $m \gets \text{\texttt{FindMinBombCovering}}(board, p)$\;
            \uIf{$m > n$ }{
                $\hat{A} \gets \hat{A} \cup \{a\}$ \;
            }
        }
    \Return $\hat{A}$ \;
  }
\caption{Action Pruning}
\label{algo:pruning}
\end{algorithm}

The essence of Algorithm~\ref{algo:pruning} is a recursive reasoning procedure that leverages the relationship between bomb's coverage and \emph{safety value}: (1) a cell is safe if it is not covered by any bombs, and (2) a bomb-covering cell is also safe if the agent has enough time and an viable way to evade to a bomb-free region.
In other words, if the board was fully observable and bombs and opponents are not moving, optimal moves will never be pruned. The immediate use of such pruning is embedding it in model-free reinforcement learning system where the goal is to learn a policy among the remaining ``safe'' actions, which we shall focus on in the remaining of the paper.  Note that Algorithm~\ref{algo:pruning} can also be used in dynamic environments by looking-ahead tree-search (i.e., by expanding all opponent moves).

Algorithm~\ref{algo:pruning} is computationally efficient, i.e., all utility functions
are computable in $O(l)$, where $l$ is width of the board, while the recursion depth of \texttt{MinEvadeStep} is less than the maximum bomb life (i.e., $10$), and the total complexity is bounded by $O(l^2)$, where $l=11$ in Pommerman.

\section{Experiments}

Here we show learning results both in modes, FFA and Team, comparing with/without action pruning.

\subsection{Learning Algorithm}
We use the PPO algorithm~\cite{schulman2017proximal} as the backend algorithm for our reinforcement learners, which minimizes the following error function on its $\mathcal{D}$, i.e., games played by the old neural net represented policy $\pi_{\theta}^{old}$:
\begin{equation} \scriptsize
\begin{split}
    L(\theta;\mathcal{D}) & = \sum_{(s_t, a_t, R_t) \in \mathcal{D}} \Bigg[ -\mathit{clip}(\frac{\pi_\theta(a_t|s_t)}{\pi_\theta^{old}(a_t|s_t)}, 1-\epsilon, 1+\epsilon) A(s_t, a_t) + \\
    & \frac{1}{4} \cdot \max\Big[ (v_\theta(s_t) -R_t)^2, (v_\theta^{old}(s_t) + \\
    & \mathit{clip}(v_\theta(s_t) - v_\theta^{old}(s_t), -\epsilon, \epsilon)-R_t)^2 \Big] \Bigg].
\end{split}
\end{equation}
Here, $R_t$ is the return at time $t$ , which can be of $-1$, $0$ or $1$ for lose, draw or win; clip range $\epsilon=0.1$, $A(s_t,a_t)$ is the generalized advantage function~\cite{schulman2015high}, where $\lambda=0.95$, discount $\gamma=0.9$, and $n\_step=32$. Those parameters were set in accordance with OpenAI baseline implementation.

\begin{figure}
    \centering
    \includegraphics[scale=0.6]{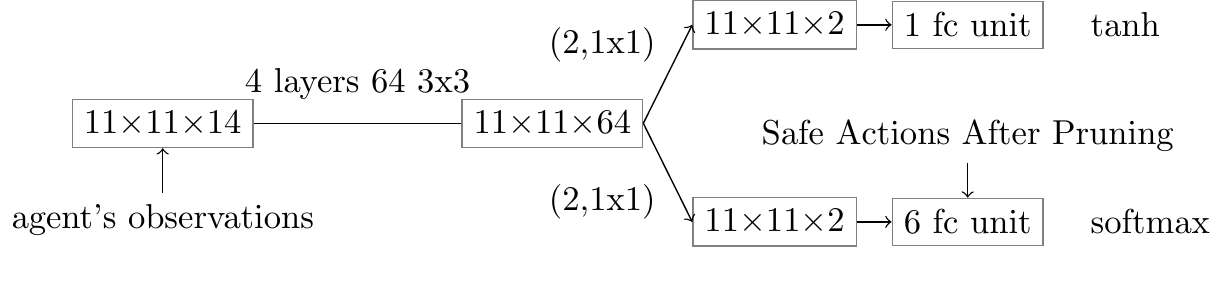}
    \caption{Architecture for learning.}
    \label{fig:arch}
\end{figure}

\subsubsection{Architecture}

Similar to previous research~\cite{resnick2018pommerman}, we extract $14$ features planes from the agent's observation which is the input to our network.
As shown in Figure~\ref{fig:arch}, the architecture contains 4 convolution layers, {followed} by two policy and value heads, respectively. The input contains 14 features planes, each of shape $11$$\times$$11$. It then convolves using $4$ layers of convolution, each has 64 $3$$\times$$3$ kernels; the result thus has shape $11$$\times$$11$$\times$$64$. Then, each head convolves using $2$ $1$$\times$$1$ kernels. Finally, the output is squashed into action probability distribution and value estimation, respectively. %

\subsubsection{Reward Shaping}
Following previous work on Pommerman~\cite{resnick2018pommerman}, to cope with the sparse reward problem, a dense reward function is added during learning:
(1) going to a cell not in a FIFO queue of size $30$ gets $0.001$ (to encourage the agent to move around {and not camp});
(2) picking up power-ups \texttt{kick, ammo, increase blast strength} gets $0.01$;
(3) killing a teammate gets \texttt{-0.5};
(4) killing an opponent gets \textsc{0.5};
and (5) blasting a piece of wood gets $0.01$.

\subsection{Learn with Static Opponents}
To show the effectiveness of our action pruning algorithm, we first learn against a team of two \emph{static} opponents, i.e., agents who alawys take the \texttt{stop} action. This reduces the problem to a task where the learning agent's role is just to remove wood, get close to, and kill opponents.

\begin{figure}
    \centering
    \begin{subfigure}{0.45\linewidth}
    \centering
    \includegraphics[scale=0.49]{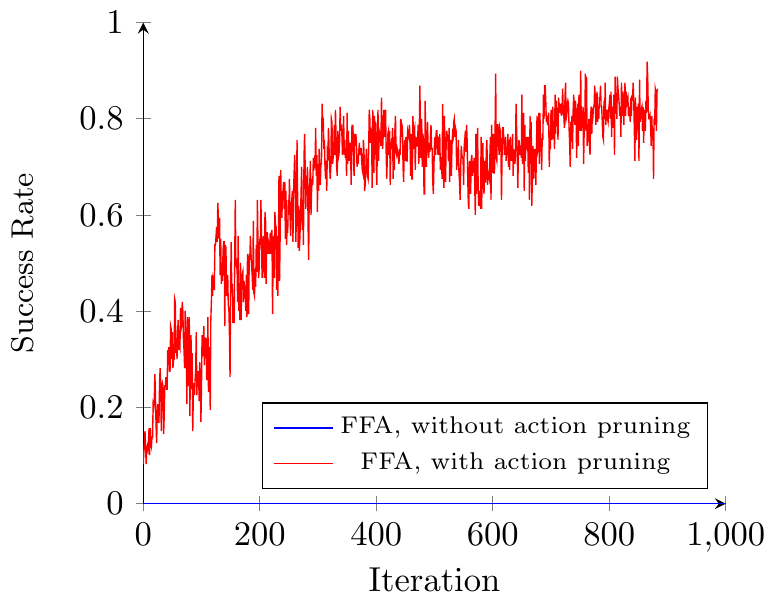}
    \\
    \caption*{(a) FFA mode}
    \end{subfigure}
    \quad
    \begin{subfigure}{0.45\linewidth}
    \centering
    \includegraphics[scale=0.49]{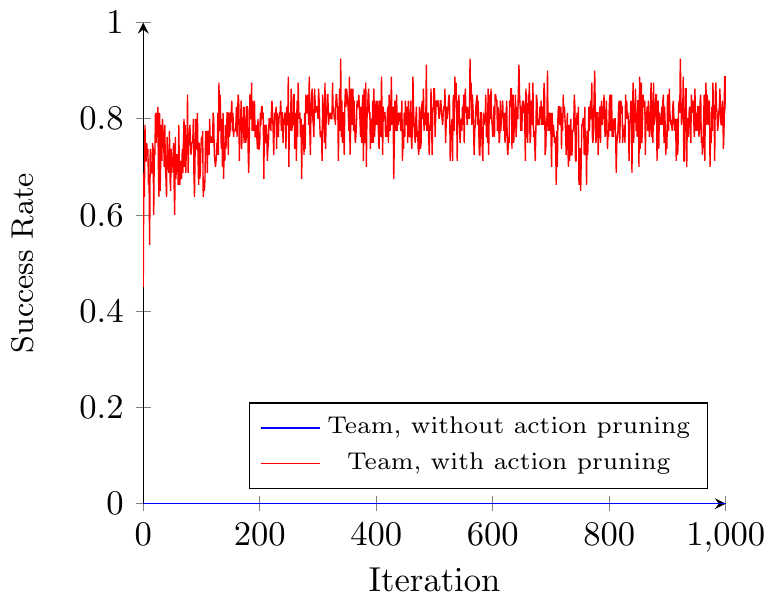}
    \\
    \caption*{(b) Team mode}
    \end{subfigure}
    \caption{Success rate against static opponents with/without action pruning module. Consistent with our prior analysis, PPO failed to learn without pruning. x-axis is the iteration number. Each iteration contains 80 games played by 8 parallel workers. When learning without pruning, the algorithm ran for 5000 iterations, but never succeeded.}
    \label{fig:static}
\end{figure}

Figure~\ref{fig:static} shows the results in both FFA and Team modes. In FFA mode, at each game, one corner agent is selected as the learning agent, and it wins if and only if it can kill the three static agents in $800$ steps. Similarly, in Team mode, two corner agents are randomly selected as neural net team players, and they win if in $800$ steps both of the two opponents are destroyed. Note that, even though static opponents may appear simple, the learning agents need to overcome obstacles (i.e, blasting wood) by bomb placement in a randomly generated board.

From Figure~\ref{fig:static}, it is clear that without any pruning, the neural nets failed to learn at all even after a training period of $5$ days --- the success rate remains zero, consistent with previous findings~\cite{resnick2018pommerman}. By contrast, after equipping the agent with the action pruning module, PPO successfully learns the tasks.  %

\subsection{Evaluate Against Baseline Opponents}
\paragraph{SimpleAgent} is a search-based baseline provided by the competition which uses hand-crafted rules for enemy interaction, bomb placement, and search for navigation.\footnote{https://github.com/multiagentlearning/playground}
\begin{table}[htp]
\footnotesize
    \centering
        \caption{Results against \texttt{SimpleAgent} in FFA and Team modes.
        In FFA, one neural net plays against three SimpleAgent; in Team, one neural net team plays against one SimpleAgent team. Final iteration neural net models on FFA and Team training are selected respectively for the test. Three independent trials were tried, where each trial contains 100 games.}%
    \begin{tabular}{c|c c c  c c c}
        \toprule
        Trial & \multicolumn{3}{c}{\texttt{FFA}}  & \multicolumn{3}{c}{\texttt{Team}} \\
         & win & draw & lose  & win & draw & lose\\
        \midrule
         1 & 43 & 12 & 45  & 69 & 6 & 25 \\
         2 & 40 & 12 & 48  & 65 & 5 & 30 \\
         3 & 42 & 17 & 41  & 76 & 6 & 18 \\
         \bottomrule
    \end{tabular}
\end{table}
To show the relative strength of our trained player, we tested the neural net model of the final iteration  against this baseline opponent. The win-percentage of our player is around 40\% against 3 \texttt{SimpleAgent}s in FFA (note that a player with similar strength of \texttt{SimpleAgent} shall win in about 25\%), and around 70\% against a \texttt{SimpleAgent} team in Team mode. These results indicate that our player outperforms \texttt{SimpleAgent} baseline, even though the neural nets were only trained against static opponents. Indeed, an early version of such a module has been used in a competition agent; by further training against a curriculum of opponents,  the final player won the second-place in the learning category at NeurIPS 2018 team competition~\cite{gao2019skynet}. The top one learning agent \texttt{Navocado} employed Dijkstra-based search, used an ensemble of several neural net models, in training and executing~\cite{peng2018continual}.

\section{Conclusions}
We have provided an analysis on the hardness of directionless exploration in Pommerman. Our analysis sheds light on the previous negative results of using model-free RL for this domain. Our analysis may also apply to other grid-world like environments for difficulty analysis. We further presented a model-based reasoning module and empirically showed its merit in both single-agent FFA and multi-agent Team environments. With the reasoning module our neural nets can learn to solve the pure exploration task given the opponents are \emph{static}, while without the reasoning module the learning agent did not improve its performance after days of training.

A number of works on analyzing the limitations of deep learning methods have emphasized on the need use \emph{models}~\cite{pearl2018theoretical,darwiche2017human}, often drawn connection to Systems I and II theories on human mind~\cite{evans2013dual}. Recent success~\cite{silver2016mastering,silver2017mastering,silver2017mastering2} in complex game playing relies on effective integration of \emph{model-based} solver and \emph{model-free} learner~\cite{geffner2018model}.
Given the exploration difficulty of Pommerman, the promising direction for high-level playing is to incorporate action-pruning (for dealing with the survival problem) and tree-search (for look-ahead opponent moves) into the multi-agent learning system.

\bibliography{aiide19}
\bibliographystyle{aaai}

\end{document}